\documentclass[10pt,twocolumn,letterpaper]{article}

\usepackage{iccv}      

\definecolor{iccvblue}{rgb}{0.21,0.49,0.74}
\usepackage[pagebackref,breaklinks,colorlinks,allcolors=iccvblue]{hyperref}

\def\paperID{*****} 
\def\confName{ICCV}
\def\confYear{2025}

\title{Self-Supervised Learning for Robotic Leaf Manipulation: A Hybrid Geometric-Neural Approach}

\author{Srecharan Selvam, \quad Abhisesh Silwal, \quad George Kantor\\
\rule{0pt}{10pt}\\
Robotics Institute, Carnegie Mellon University\\
{\tt\small \{sselvam, asilwal, gkantor\}@andrew.cmu.edu}
}

\begin{document}
\maketitle
\begin{abstract}
\it{Automating leaf manipulation in agricultural settings faces significant challenges, including the variability of plant morphologies and deformable leaves. We propose a novel hybrid geometric-neural approach for autonomous leaf grasping that combines classical computer vision with neural networks through self-supervised learning. Our method integrates YOLOv8 for instance segmentation and RAFT-Stereo for 3D depth estimation to build rich leaf representations, which feed into both a geometric feature scoring pipeline and a neural refinement module (GraspPointCNN). The key innovation is our confidence-weighted fusion mechanism that dynamically balances the contribution of each approach based on prediction certainty. Our self-supervised framework uses the geometric pipeline as an expert teacher to automatically generate training data. Experiments demonstrate that our approach achieves an 88.0\% success rate in controlled environments and 84.7\% in real greenhouse conditions, significantly outperforming both purely geometric (75.3\%) and neural (60.2\%) methods. This work establishes a new paradigm for agricultural robotics where domain expertise is seamlessly integrated with machine learning capabilities, providing a foundation for fully automated crop monitoring systems.}
\end{abstract}    
\section{Introduction}

Agricultural robotics has emerged as a key technology for addressing labor shortages and improving efficiency in modern farming \cite{Bechar2016Agricultural,Shamut2021AgriLabor}. Among greenhouse cultivation tasks, leaf sampling for disease detection remains a significant bottleneck, requiring skilled workers to manually identify, select, and extract tissue samples from thousands of plants \cite{Shamshiri2018Research,Atefi2021PlantPath}. This labor-intensive process increases operational costs and limits the frequency of plant health monitoring, potentially allowing diseases to spread undetected \cite{Bac2014Harvesting,Lu2020EarlyDetection}.

Automating leaf manipulation presents unique challenges compared to traditional robotic grasping tasks. Unlike rigid industrial objects, plant leaves are deformable, vary significantly in size and orientation, and are often partially occluded in dense canopies \cite{Lehnert2017Autonomous,Nguyen2018DeformGrasp}. While recent advances in deep learning have revolutionized robotic grasping for industrial applications \cite{Mahler2017DexNet,Morrison2018Closing,Zeng2019GraspingSurvey}, these approaches typically require large datasets of labeled grasp points—a resource that is prohibitively expensive to create for agricultural settings where plant morphology varies continuously throughout growth cycles \cite{Koirala2019DeepLearningAgri}.

Existing approaches to agricultural manipulation fall into two categories: purely geometric methods that rely on hand-crafted features \cite{Hemming2014Fruit,Silwal2017Design,Hughes2021GeometricGrasping}, and end-to-end deep learning systems trained on synthetic or limited real-world data \cite{Arad2020Development,Yu2019Fruit,Wang2021DataAugmentation}. Geometric approaches, while interpretable and robust to domain shifts, struggle with the natural variability of plant structures. Conversely, deep learning methods excel at handling complex visual patterns but suffer from poor generalization when deployed on new crop varieties or growth stages not represented in their training data \cite{Milioto2018CrossDomain}.

We present a novel hybrid approach that leverages the complementary strengths of geometric reasoning and neural networks through self-supervised learning. Our key insight is that traditional computer vision algorithms, despite their limitations, encode valuable domain expertise that can serve as a teacher for training neural networks without manual annotation \cite{Jha2021SelfSupervision}. This approach enables continuous learning from operational data while maintaining the interpretability and reliability required for agricultural automation.

Our system operates on a 6-DOF gantry robot equipped with stereo vision and a custom end-effector for leaf manipulation. The perception pipeline combines YOLOv8 instance segmentation \cite{Jocher2023YOLO} with RAFT-Stereo depth estimation \cite{Lipson2021RAFTStereo} to generate 3D representations of plant canopies. For grasp point selection, we implement a dual-path architecture: a geometric pipeline using Pareto optimization across multiple hand-crafted features (flatness, accessibility, edge distance), and a convolutional neural network with spatial attention that learns from the geometric system's decisions \cite{Kendall2018MultiTaskUncertainty}.

The main contributions of this work include:
\begin{itemize}
\item A self-supervised learning framework where geometric algorithms act as expert teachers for neural networks, eliminating the need for manual grasp annotation in agricultural settings
\item A hybrid decision architecture that dynamically weighs geometric and learned features based on prediction confidence, achieving robust performance across diverse plant conditions
\item A comprehensive grasp point selection system incorporating novel scoring functions tailored to leaf-specific constraints such as deformability, approach angles, and occlusion handling
\item Extensive validation on thousands of real plant samples demonstrating significant improvements over traditional geometric methods, particularly in challenging scenarios with partial occlusion and irregular orientations
\end{itemize}
This work provides a foundation for fully automated crop monitoring systems and establishes a new paradigm for agricultural robotics where domain expertise is seamlessly integrated with machine learning capabilities.
\section{Related Work}

\subsection{Vision-Based Leaf Manipulation}

Traditional approaches to robotic leaf manipulation in agricultural settings relied on geometric reasoning and classical computer vision. Hemming et al. developed methods for cucumber leaf detection in greenhouses using color and texture features \cite{Hemming2014Robot}, while Bac et al. presented obstacle-aware motion planning for tomato canopies \cite{Bac2017Performance}. Several studies focused on deformable leaf modeling, including Cerutti et al.'s parametric active polygon models \cite{Cerutti2013Parametric}, Xia et al.'s active shape models for overlapping leaves \cite{Xia2018Plant}, and Jin et al.'s probabilistic graphical models for leaf structure analysis \cite{Jin2018PGMLeaf}. The integration of 3D information improved robustness, as demonstrated by Guo and Xu's multiview stereo reconstruction for lettuce segmentation \cite{Guo2017Leaf} and Sodhi et al.'s plant growth monitoring system using structure-from-motion \cite{Sodhi2020PhenotypingSystem}. While effective in controlled conditions, these methods often required extensive tuning and struggled with natural plant variability, particularly under varying illumination conditions \cite{Dandrifosse2021RobustTracking}.

\subsection{Deep Learning for Agricultural Grasping}

Deep learning has shown promise in agricultural manipulation, though with unique challenges compared to industrial applications. Barth et al. developed CNN-based systems for broccoli harvesting that handle significant occlusion \cite{Barth2019Synthetic}, while Arad et al. demonstrated sweet pepper harvesting combining YOLO detection with stereo depth \cite{Arad2020Development}. For leaf-specific tasks, Ahlin et al. pioneered CNN-based leaf identification with visual servoing for autonomous sampling, achieving 85\% success rates in greenhouses \cite{Ahlin2016Autonomous}. However, these approaches typically require extensive training data—a significant limitation given the continuous variation in plant morphology \cite{Ubbens2020DataAugmentation}. To address this, researchers have explored simulation, with approaches like Dex-Net generating synthetic grasp scenarios \cite{Mahler2017DexNet}, inspiring agricultural adaptations for data generation \cite{Kuznetsova2020TransferLearningAgri}. Recent advances in generative data augmentation have shown promise for bridging the synthetic-real domain gap \cite{Valada2020DomainAdaptation}, particularly for multi-crop environments with challenging lighting conditions.

\subsection{Self-Supervised Learning in Agricultural Robotics}

Self-supervised learning has emerged as a promising paradigm for agricultural robotics, particularly where manual annotation is expensive. Zhang et al. demonstrated self-supervised learning for tomato harvesting, using classical vision systems to provide training labels \cite{Zhang2021Selfsupervised}. Similar bootstrapping approaches include Kootstra et al.'s work on sweet pepper detection, where geometric algorithms generated training data for neural networks \cite{Kootstra2021Selective} and Tao et al.'s approach using temporal consistency for plant growth tracking \cite{Tao2022TemporalConsistency}. This knowledge transfer from classical to learning-based systems has proven particularly valuable in controlled environment agriculture, where hybrid approaches consistently outperform purely learned policies \cite{Shamshiri2018Research,Garrido2019SelfSupervisedAgri}. Recent work has also explored contrastive learning frameworks for agricultural visual representations \cite{Weyand2021ContrastiveAgri}, enabling more sample-efficient adaptation to new crops and growth conditions.

\subsection{3D Perception and Hybrid Systems}
\begin{figure*}[t]
    \centering
    \includegraphics[width=\textwidth]{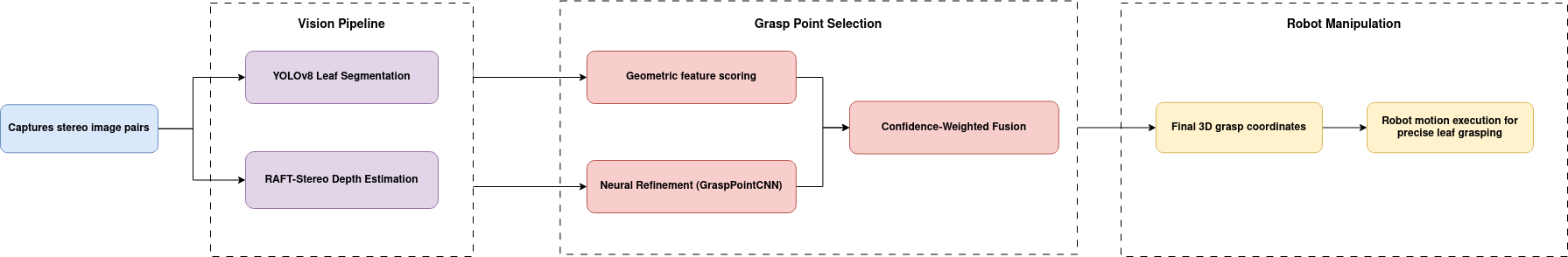}
    \caption{System architecture showing the integration of vision pipeline, grasp point selection, and robot manipulation modules. The hybrid approach combines geometric feature scoring with neural refinement through confidence-weighted fusion.}
    \label{fig:system_pipeline}
\end{figure*}

Accurate depth sensing is crucial for manipulation in dense plant canopies. While traditional stereo algorithms struggle with plant textures, recent advances like RAFT-Stereo have dramatically improved accuracy for agricultural applications \cite{Lipson2021RAFTStereo}. Lipson et al.'s recurrent architecture achieves state-of-the-art performance on challenging plant datasets, enabling precise leaf pose estimation \cite{Sa2017Peduncle}. Alternative sensing modalities such as time-of-flight cameras \cite{McCool2019ToFSensing} and structured light systems \cite{Zhou2019StructuredLightPlants} have also shown promise for plant phenotyping applications with complex geometries.
Recent research increasingly combines classical and learning approaches, as demonstrated by Lehnert et al.'s hybrid system for pepper harvesting \cite{Lehnert2016Sweet} and Adamides et al.'s framework for integrating human expertise with machine learning \cite{Adamides2021HybridDecision}. These hybrid architectures leverage geometric interpretability with neural adaptability, making them ideal for complex agricultural tasks where safety and reliability are paramount \cite{Duckett2018Agricultural}. Confidence-aware systems that adaptively balance multiple decision sources have proven especially effective in scenarios with high uncertainty, demonstrating robustness across varying environmental conditions \cite{Gao2020ConfidenceAware}.

\section{Method}

We present a hybrid approach for autonomous leaf grasping that combines geometric algorithms with neural networks through self-supervised learning. Our system eliminates the need for manual grasp annotation while maintaining robust performance in complex greenhouse environments. This section details our perception pipeline, grasp point selection algorithms, and the self-supervised framework that bridges classical and modern approaches.

\subsection{System Overview}

Figure~\ref{fig:system_pipeline} presents our hybrid leaf grasping system architecture, consisting of three modules: vision pipeline, grasp point selection, and robot manipulation. The system processes stereo image pairs from a 6-DOF gantry robot to output precise 3D grasp coordinates.

The vision pipeline employs YOLOv8 for instance segmentation of individual leaves and RAFT-Stereo for dense depth estimation. As shown in Figure~\ref{fig:system_pipeline}, these outputs are fused to create 3D leaf representations containing both semantic and geometric information.

The grasp point selection module implements our hybrid approach through two parallel paths. The geometric feature scoring path evaluates candidates using traditional CV algorithms based on features like flatness, accessibility, and approach angles. Simultaneously, the neural refinement path (GraspPointCNN) processes the same data using learned features. Both predictions are combined through confidence-weighted fusion, dynamically balancing traditional CV (70-90\%) and neural network (10-30\%) contributions.

Our key innovation is the self-supervised training scheme where geometric algorithms act as expert teachers, automatically labeling grasp points to train the neural network. This enables the system to initially mimic geometric reasoning while developing generalization capabilities beyond hand-crafted features.

The robot manipulation module executes precise leaf grasping using the final 3D coordinates, with motion planning optimized for the gantry configuration and safety validation through force feedback.

\subsection{Vision Pipeline}
The vision pipeline, illustrated in the left section of Figure~\ref{fig:system_pipeline}, processes stereo image pairs to generate rich 3D representations of plant leaves. This pipeline employs two parallel processing streams: instance segmentation and stereo depth estimation, whose outputs are fused to create comprehensive leaf models for grasp planning.

\subsubsection{Instance Segmentation}

We utilize YOLOv8 \cite{Jocher2023YOLO} for real-time instance segmentation of individual leaves. Unlike standard implementations, we fine-tuned YOLOv8 on a custom dataset of 900+ images containing soybean and tomato plants in greenhouse conditions. This domain-specific training enables robust leaf detection even in challenging scenarios with significant overlap and occlusion, achieving 90\%+ confidence scores in operational conditions.

\begin{figure}[t]
    \centering
    \includegraphics[width=\columnwidth]{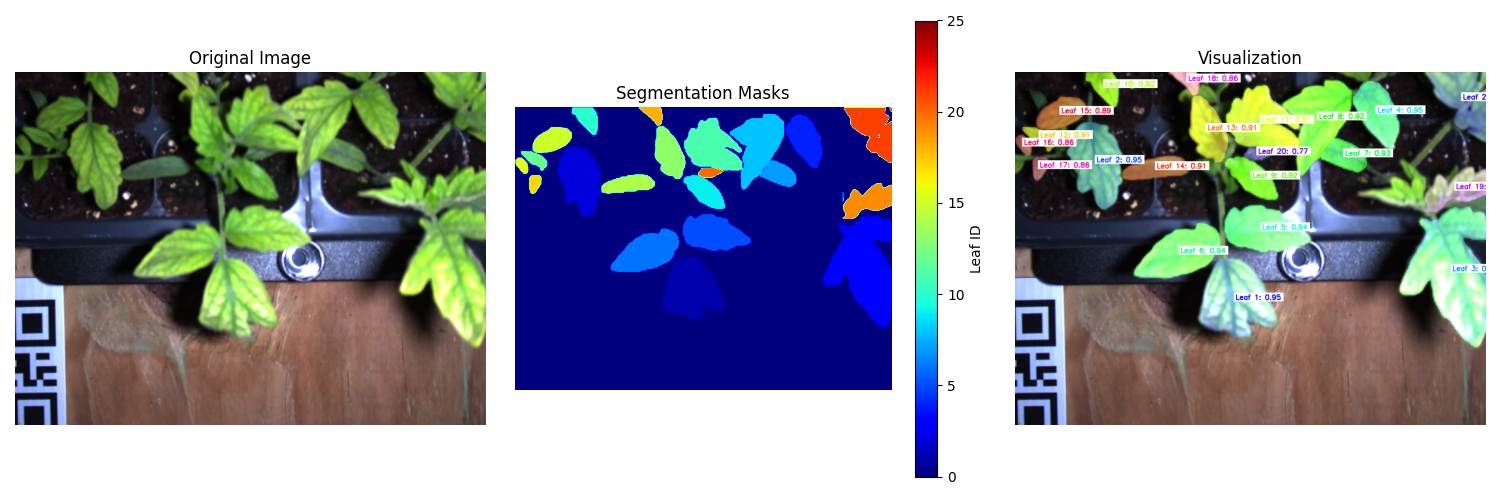}
    \caption{Vision pipeline outputs: (a) Instance segmentation with individual leaf masks, (b) RAFT-Stereo disparity map, (c) 3D point cloud reconstruction with highlighted target leaf.}
    \label{fig:yolo_segmentation}
\end{figure}

As shown in Figure~\ref{fig:yolo_segmentation}, the network outputs binary masks for each detected leaf instance along with confidence scores. The segmentation accurately delineates individual leaf boundaries despite complex overlapping patterns typical in dense canopies. Our implementation processes 1440×1080 resolution images at approximately 50ms per frame, meeting real-time requirements for robotic manipulation. Each detected leaf is assigned a unique identifier and confidence score, enabling robust tracking throughout the grasp selection process.

\subsubsection{Stereo Depth Estimation}

For 3D reconstruction, we employ RAFT-Stereo \cite{Lipson2021RAFTStereo}, which generates dense disparity maps through iterative refinement using recurrent all-pairs field transforms. This approach handles the thin structures and low-texture regions characteristic of plant foliage more reliably than traditional stereo matching algorithms \cite{Scharstein2002Taxonomy}. 

\begin{figure}[t]
    \centering
    \begin{tabular}{cc}
        \includegraphics[width=0.48\columnwidth]{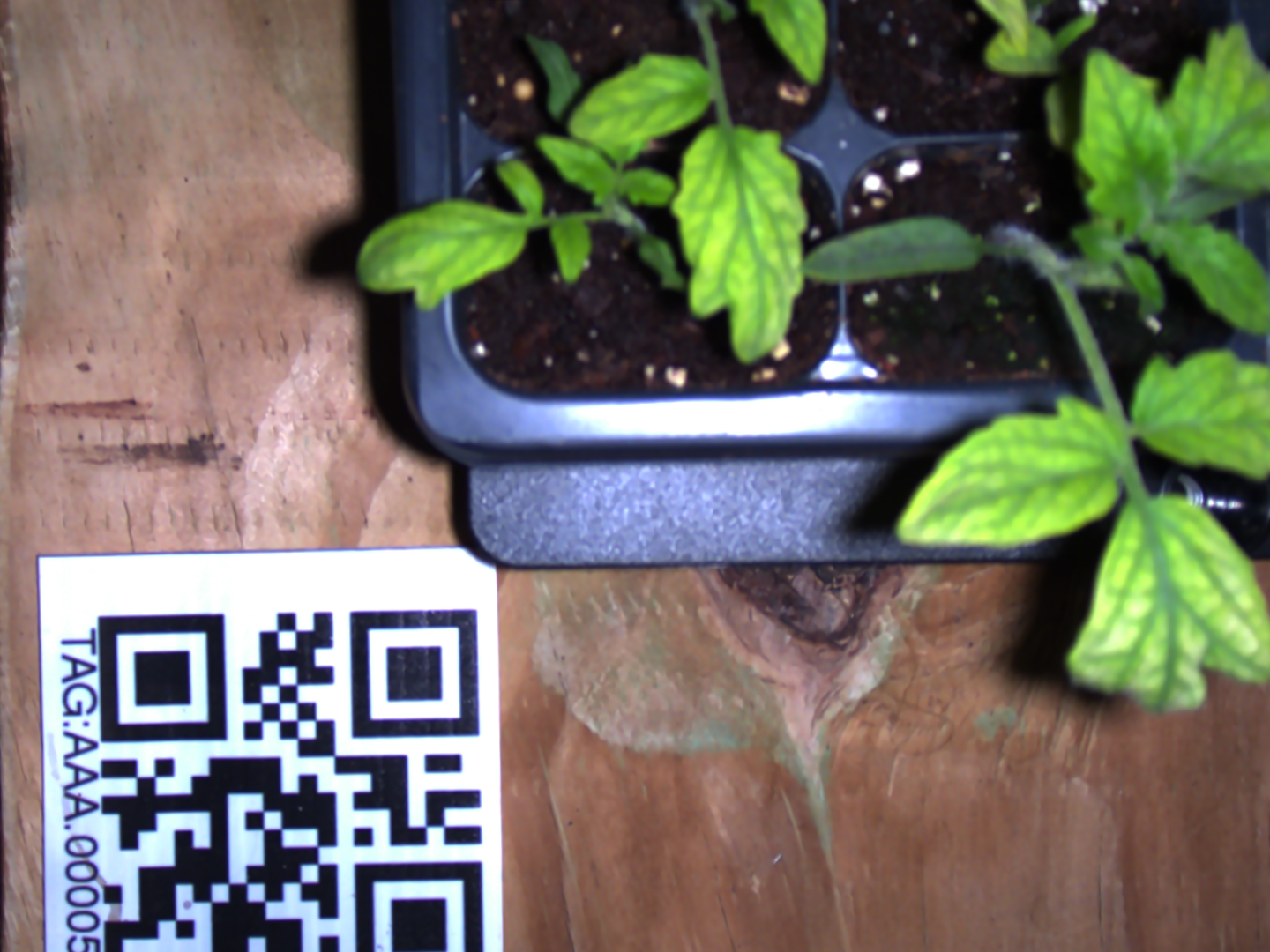} & 
        \includegraphics[width=0.48\columnwidth]{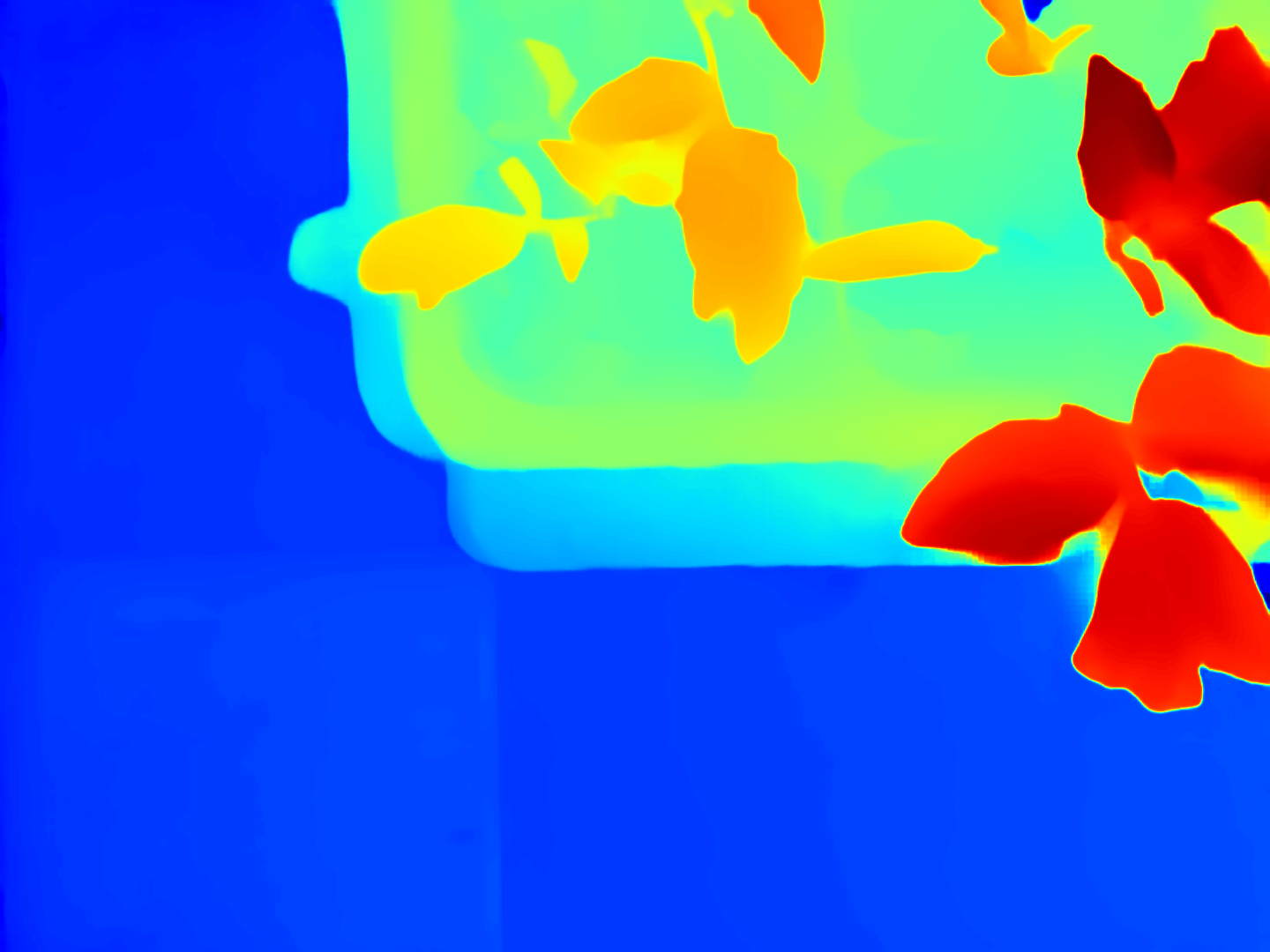} \\
        (a) Raw stereo image & (b) RAFT-Stereo depth map \\
        \multicolumn{2}{c}{\includegraphics[width=0.98\columnwidth]{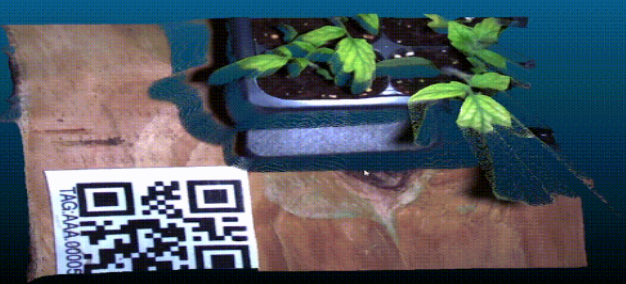}} \\
        \multicolumn{2}{c}{(c) 3D leaf reconstruction from stereo depth and segmentation}
    \end{tabular}
    \caption{RAFT-Stereo outputs showing the processing pipeline: (a) Raw image from the left camera of the stereo pair, (b) Generated disparity map where warmer colors indicate closer objects, (c) Final 3D reconstruction combining depth and segmentation data.}
    \label{fig:raft_stereo}
\end{figure}

Our calibrated stereo pair captures synchronized images at 1440$\times$1080 resolution. As illustrated in Figure~\ref{fig:raft_stereo}, RAFT-Stereo processes these to produce sub-pixel accurate disparity maps in approximately 60ms, achieving 29\% lower 1-pixel error than previous methods on standard benchmarks \cite{Geiger2012KITTI}. 
The disparity values are converted to metric depth using the camera calibration parameters, enabling accurate 3D reconstruction.

\subsubsection{3D Reconstruction}
Each pixel $(u,v)$ with disparity $d$ is back-projected to 3D coordinates $(X,Y,Z)$ using:
\begin{equation}
X = \frac{(u - c_x) \cdot Z}{f_x}, \quad Y = \frac{(v - c_y) \cdot Z}{f_y}, \quad Z = \frac{f \cdot b}{d}
\end{equation}
where $f$ is the focal length, $b$ is the stereo baseline, and $(c_x, c_y)$ are the principal point coordinates. The resulting point cloud provides comprehensive 3D structure of the scene, as shown in Figure~\ref{fig:raft_stereo}(c).

\subsubsection{Data Fusion}

The vision pipeline combines segmentation masks with depth information to create per-leaf 3D models. For each detected leaf instance, we:
\begin{itemize}
    \item Extract 3D points by masking the depth map with the leaf's segmentation mask
    \item Compute geometric properties including centroid position, surface area, and orientation
    \item Estimate surface normals through local plane fitting for flatness evaluation
    \item Identify occlusion by detecting missing depth data within mask boundaries
\end{itemize}

This fusion process outputs a structured representation of each leaf containing both 2D mask information and 3D geometric properties, providing the necessary data for subsequent grasp point selection algorithms. The geometric processing includes signed distance field (SDF) generation, which will be detailed in Section~\ref{sec:traditional_cv}.

\subsection{Geometric Feature Scoring Pipeline}
\label{sec:traditional_cv}

The geometric feature scoring pipeline evaluates candidate leaves and grasp points using hand-crafted features derived from classical computer vision principles. This deterministic approach provides interpretable decisions and serves as the foundation for our self-supervised learning framework.

\subsubsection{Optimal Leaf Selection}

Given the set of segmented leaves from the vision pipeline, we evaluate each leaf using three key metrics: clutter, distance, and visibility. These metrics are combined using Pareto optimization to identify the optimal grasping target.

\begin{equation}
L^* = \underset{L_i \in \mathcal{L}}{\arg\max} \Big( w_c S_c(L_i) + w_d S_d(L_i) + w_v S_v(L_i) \Big)
\end{equation}

Where:
\begin{itemize}
    \item $L^*$ is the optimal leaf selection
    \item $\mathcal{L}$ is the set of all detected leaves  
    \item $S_c(L_i)$ is the clutter/isolation score for leaf $i$
    \item $S_d(L_i)$ is the distance score for leaf $i$
    \item $S_v(L_i)$ is the visibility score for leaf $i$
    \item $w_c, w_d, w_v$ are the weights (0.35, 0.35, 0.30)
\end{itemize}

\textbf{Clutter Score} quantifies leaf isolation using signed distance fields (SDF):

\begin{equation}
S_{clutter} = \frac{d_{min}}{d_{min} + d_{max}}
\end{equation}

Where:
\begin{itemize}
    \item $d_{min}$ is the distance from centroid to SDF minimum
    \item $d_{max}$ is the distance from centroid to SDF maximum
\end{itemize}

\textbf{Distance Score} evaluates the leaf's 3D Euclidean distance from the camera:

\begin{equation}
S_{distance} = e^{-\frac{d_{mean}}{0.3}}
\end{equation}

Where:
\begin{itemize}
    \item $d_{mean}$ is the mean Euclidean distance of leaf points
    \item 0.3m is the scale factor
\end{itemize}

\textbf{Visibility Score} assesses leaf completeness and position:

\begin{equation}
S_{visibility} = 
\begin{cases}
0 & \text{if leaf touches image border} \\
1 - \frac{d_{center}}{d_{max}} & \text{otherwise}
\end{cases}
\end{equation}

Where:
\begin{itemize}
    \item $d_{center}$ is the distance from leaf centroid to image center
    \item $d_{max}$ is the maximum possible distance in the image
\end{itemize}

The final leaf selection employs Pareto optimization with weighted scoring:

\begin{equation}
S_{leaf} = 0.35 \cdot S_{clutter} + 0.35 \cdot S_{distance} + 0.30 \cdot S_{visibility}
\end{equation}

\begin{figure}[t]
    \centering
    \begin{tabular}{cc}
        \includegraphics[width=0.48\columnwidth]{figure/raw_image.png} & 
        \includegraphics[width=0.48\columnwidth]{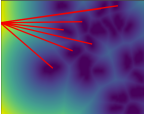} \\
        (a) Raw image with candidates & (b) SDF representation
    \end{tabular}
    \caption{Signed Distance Field (SDF) visualization for grasp planning: (a) Raw plant image with leaf candidates, (b) SDF representation showing free space (purple/blue) and occupied regions (yellow/red). Red rays indicate potential grasp approach directions.}
    \label{fig:sdf_visualization}
\end{figure}

Figure~\ref{fig:sdf_visualization} illustrates the SDF computation used for clutter evaluation. The SDF representation enables efficient calculation of clearance around each leaf candidate, with warmer colors indicating proximity to obstacles.

\subsubsection{Geometric Grasp Point Scoring}

Once the target leaf is selected, we generate candidate grasp points uniformly distributed across the leaf surface. Each candidate is evaluated using four geometric features:

\begin{equation}
\begin{aligned}
G^* = \underset{p \in L^*}{\arg\max} \Big( & w_f F(p) + w_a A(p) + \\
& w_e E(p) + w_{acc} Acc(p) \Big) \cdot (1 - S_{pen}(p))
\end{aligned}
\end{equation}

Where:
\begin{itemize}
    \item $G^*$ is the optimal grasp point
    \item $p$ is a candidate point on the selected leaf $L^*$
    \item $F(p)$ is the flatness score at point $p$
    \item $A(p)$ is the approach vector alignment score at point $p$
    \item $E(p)$ is the edge margin score at point $p$
    \item $Acc(p)$ is the accessibility score at point $p$
    \item $S_{pen}(p)$ is the stem penalty term
    \item $w_f, w_a, w_e, w_{acc}$ are the weights (0.25, 0.40, 0.20, 0.15)
\end{itemize}

\textbf{Flatness Score} measures local surface planarity using depth gradients:

\begin{equation}
F(p) = e^{-\alpha \cdot \sqrt{|\nabla_x D(p)|^2 + |\nabla_y D(p)|^2}}
\end{equation}

Where:
\begin{itemize}
    \item $D(p)$ is the depth value at point $p$
    \item $\nabla_x D$ and $\nabla_y D$ are the gradients in x and y directions
    \item $\alpha = 5.0$ is the scaling factor
\end{itemize}

\textbf{Approach Vector Alignment} evaluates grasp accessibility:

\begin{equation}
A(p) = \Big|\frac{\vec{v}(p) \cdot \vec{z}}{|\vec{v}(p)|}\Big|
\end{equation}

Where:
\begin{itemize}
    \item $\vec{v}(p)$ is the vector from camera to point $p$
    \item $\vec{z}$ is the unit vector in the vertical direction (0,0,1)
\end{itemize}

\textbf{Edge Distance Score} penalizes points near leaf boundaries:

\begin{equation}
E(p) = \min\Big(1, \frac{d_{edge}(p)}{d_{safe}}\Big)
\end{equation}

Where:
\begin{itemize}
    \item $d_{edge}(p)$ is the distance to the nearest edge
    \item $d_{safe} = 5$mm is the minimum safe distance
\end{itemize}

\textbf{Accessibility Score} considers kinematic reachability:

\begin{equation}
Acc(p) = 0.7 \cdot \Big(1 - \frac{d(p, c)}{d_{max}}\Big) + 0.3 \cdot \cos(\theta(p))
\end{equation}

Where:
\begin{itemize}
    \item $d(p, c)$ is the distance from point $p$ to the image center
    \item $d_{max}$ is the maximum distance in the image
    \item $\theta(p)$ is the angle between the vector to point $p$ and the forward direction
\end{itemize}

The final grasp quality score combines these metrics:

\begin{equation}
S_{grasp} = 0.25 \cdot F(p) + 0.40 \cdot A(p) + 0.20 \cdot E(p) + 0.15 \cdot Acc(p)
\end{equation}

\begin{figure}[t]
    \centering
    \begin{minipage}{\columnwidth}
        \centering
        \includegraphics[width=0.95\columnwidth]{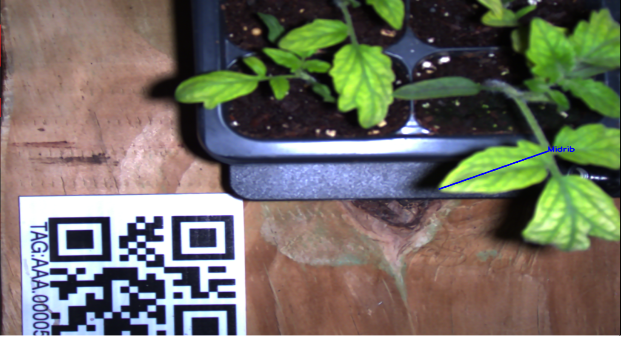}
        \caption*{(a) Raw input image from the stereo camera}
    \end{minipage}
    
    \vspace{0.5cm}
    
    \begin{minipage}{\columnwidth}
        \centering
        \includegraphics[width=0.95\columnwidth]{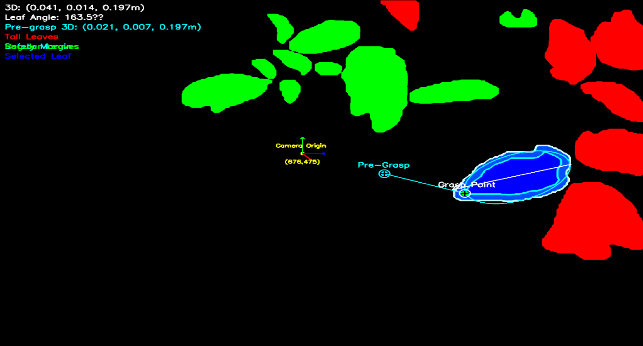}
        \caption*{(b) Geometric Feature Scoring Pipeline output}
    \end{minipage}
    
    \caption{Grasp point selection visualization. (a) Raw camera image showing leafs and optimal leaf's midrib. (b) Geometric feature scoring output showing selected leaf (blue outline), candidate grasp points, and final selected grasp point with approach vector. The visualization includes safety margins and coordinate information.}
    \label{fig:geometric_output}
\end{figure}

Figure~\ref{fig:geometric_output} demonstrates the complete geometric pipeline output, showing the selected leaf, evaluated grasp candidates, and the final chosen grasp point with its 3D coordinates. This deterministic output serves as ground truth for training our neural refinement module, detailed in the following section.

\subsubsection{Stem Proximity Penalty}

An additional penalty is applied to prevent grasping near the leaf stem:

\begin{equation}
S_{final} = S_{grasp} \cdot (1 - S_{stem\_penalty})
\end{equation}

Where:
\begin{itemize}
    \item $S_{stem\_penalty} = e^{-\alpha \cdot d_{stem}}$
    \item $d_{stem}$ is the distance to the detected stem region
    \item $\alpha = 0.1$ is the decay factor
\end{itemize}

The geometric pipeline outputs a grasp proposal consisting of the selected leaf index and optimal grasp point coordinates, providing a robust baseline for our hybrid system.

Despite its effectiveness, the geometric pipeline has several limitations. It struggles with irregular leaf morphologies not captured by hand-crafted features, requires extensive parameter tuning across plant species, and performs inconsistently in scenarios with dense occlusion or unusual lighting conditions. The correlation coefficients between expert-selected grasp points and geometric pipeline selections drop significantly from 0.92 for ideal conditions to 0.68 for challenging scenarios. These limitations motivate our neural refinement module (GraspPointCNN), which learns from the geometric system's successes while developing generalization capabilities beyond hand-crafted features, particularly for edge cases where traditional computer vision approaches falter.

\subsection{Neural Refinement Module (GraspPointCNN)}
\label{sec:grasp_point_cnn}

While the geometric feature scoring pipeline provides a robust baseline for leaf grasping, its fixed heuristics limit adaptability to novel plant morphologies and environmental conditions. We introduce GraspPointCNN, a convolutional neural network with spatial attention that learns to evaluate grasp candidates by capturing complex patterns beyond hand-crafted features.

\subsubsection{Network Architecture}

GraspPointCNN employs a compact yet effective architecture designed for real-time inference. The network consists of:

\noindent\textbf{Input Layer:} A 9-channel feature representation combining:
\begin{itemize}
    \item Depth patch (1 channel): Local 3D structure information
    \item Binary segmentation mask (1 channel): Leaf boundary information
    \item Geometric score maps (7 channels): Individual component scores from the traditional pipeline
\end{itemize}

\noindent\textbf{Encoder Blocks:} Three sequential encoder blocks, each containing:
\begin{itemize}
    \item 2D convolution (kernel size 3$\times$3, stride 1)
    \item Batch normalization
    \item ReLU activation
    \item Max pooling (2$\times$2, stride 2)
\end{itemize}
The three-encoder architecture provides an optimal balance between computational efficiency and feature extraction capacity, as determined through ablation studies comparing 2-5 encoder variants.

\noindent\textbf{Spatial Attention Mechanism:} A novel leaf-specific attention module that emphasizes salient regions:
\begin{equation}
\label{eq:spatial_attention}
\begin{aligned}
F_{spatial} &= \sigma(\text{Conv}_{7\times7}(\text{Concat}[AvgPool(F), MaxPool(F)])) \\
F_{att} &= F \odot F_{spatial}
\end{aligned}
\end{equation}
Where:
\begin{itemize}
    \item $F$ represents feature maps
    \item $\sigma$ is the sigmoid activation
    \item $\odot$ denotes element-wise multiplication
\end{itemize}
This attention mechanism allows the network to focus on critical leaf features such as venation patterns, curvature transitions, and surface variations that impact graspability.

\noindent\textbf{Decision Layers:} The network concludes with:
\begin{itemize}
    \item Global average pooling to ensure translation invariance
    \item Two fully-connected layers (128 and 64 neurons)
    \item Sigmoid activation producing a final grasp quality score [0,1]
\end{itemize}

The compact design (approximately 285K parameters) enables inference in under 10ms on standard GPU hardware, making it suitable for real-time robotic applications.

\subsubsection{Input Representation}
\label{sssec:input_representation}

For each candidate grasp point, we extract a 32$\times$32 pixel patch centered at the point from the following sources:
\begin{equation}
\label{eq:input_representation}
X_{input} = [X_{depth}, X_{mask}, X_{scores}]
\end{equation}
Where:
\begin{itemize}
    \item $X_{depth}$ is the normalized local depth patch
    \item $X_{mask}$ is the binary segmentation mask
    \item $X_{scores}$ contains seven geometric score maps (flatness, approach vector, edge distance, accessibility, etc.)
\end{itemize}
This multi-modal representation combines geometric, semantic, and raw depth information, enabling the network to reason about both local and contextual factors affecting grasp success. By incorporating the individual component scores from the traditional pipeline, the network can learn which features are most relevant in different scenarios, effectively developing an adaptive weighting scheme.

\subsubsection{Confidence Estimation}
\label{sssec:confidence_estimation}

A key innovation in our approach is the estimation of prediction confidence alongside grasp quality scores. Rather than simply outputting a binary classification, GraspPointCNN produces a continuous score that encodes both grasp quality and prediction certainty:
\begin{equation}
\label{eq:confidence_estimation}
C_{pred} = 1.0 - \left| S_{pred} - 0.5 \right| \times 2
\end{equation}
Where:
\begin{itemize}
    \item $S_{pred}$ is the raw network output [0,1]
    \item $C_{pred}$ is the confidence score [0,1]
\end{itemize}
This formulation yields maximum confidence (1.0) for extreme predictions (0 or 1) and minimum confidence (0) for uncertain predictions (0.5). The confidence estimation enables our hybrid integration system to dynamically balance traditional and learned approaches based on prediction reliability.

The neural architecture effectively addresses the limitations of pure geometric approaches through:
\begin{itemize}
    \item Generalization to novel morphologies: By learning from diverse leaf examples, the network generalizes to plant structures not explicitly encoded in hand-crafted features
    \item Contextual understanding: The spatial attention mechanism captures relationships between local surface properties and broader leaf context
    \item Adaptability to environmental variations: Learning from operational data across different lighting conditions and growth stages enables robustness to environmental changes
    \item Uncertainty awareness: The confidence estimation provides critical information for safe hybrid decision-making
\end{itemize}
The GraspPointCNN complements the geometric pipeline by focusing on capturing patterns that emerge from complex interactions between multiple factors, rather than treating each feature independently. This holistic approach is particularly valuable for edge cases where traditional CV approaches falter.

\subsection{Self-Supervised Learning Framework}
\label{sec:self_supervised_learning_v2}

A key challenge in developing learning-based robotic grasp systems for agriculture is the lack of labeled training data. We address this through a self-supervised framework where the geometric pipeline acts as an expert teacher, automatically generating training data without human intervention.

\subsubsection{Automatic Training Data Generation}
\label{sssec:automatic_data_generation_v2}

Our approach leverages the geometric pipeline to create a continuously growing dataset:

\begin{enumerate}
    \item \textbf{Positive Sample Collection}: During operation, the system captures successful grasp points selected by the geometric pipeline along with their local context (32$\times$32 pixel patches).

    \item \textbf{Data Augmentation}: To increase sample diversity, we employ:
    \begin{itemize}
        \item Rotational transformations (90\textdegree, 180\textdegree, 270\textdegree)
        \item Random cropping with 0.9-1.0 scale factor
        \item Mild brightness and contrast adjustments ($\pm$10\%)
        \item Gaussian noise injection ($\sigma = 0.01$)
        \item Random horizontal flipping
    \end{itemize}

    \item \textbf{Negative Sample Generation}: We systematically identify challenging regions:
    \begin{itemize}
        \item Leaf tips (distance transform maxima)
        \item Stem regions (morphological analysis)
        \item High-curvature edges (depth gradient thresholding)
    \end{itemize}

    \item \textbf{Validation Filtering}: An automated quality assessment removes low-quality samples based on depth completion, segmentation quality, and score consistency.
\end{enumerate}

This process yielded a dataset with the following composition:

\begin{table}[t]
\centering
\begin{tabular}{lr}
\hline
\textbf{Dataset Component} & \textbf{Count} \\
\hline
Original Positive Samples & 125 \\
Augmented Positive Samples & 375 \\
Negative Samples & 375 \\
\hline
\textbf{Total Dataset Size} & \textbf{875} \\
\hline
\end{tabular}
\caption{Composition of the self-supervised training dataset.}
\label{tab:self_supervised_dataset_v2}
\end{table}

\subsubsection{Training Methodology}
\label{sssec:training_methodology_v2}

GraspPointCNN was trained using binary cross-entropy loss with positive class weighting:

\begin{equation}
\label{eq:loss_function_v2}
\mathcal{L} = -\frac{1}{N}\sum_{i=1}^{N} [w_p \cdot y_i \log(\hat{y}_i) + (1-y_i)\log(1-\hat{y}_i)]
\end{equation}
Where $y_i$ is the ground truth label, $\hat{y}_i$ is the predicted score, and $w_p = 2.0$ is the positive class weight.

The model was trained with:
\begin{itemize}
    \item Learning rate: 0.0005
    \item Weight decay: 0.01
    \item Batch size: 16
    \item Early stopping: 15 epochs patience
\end{itemize}

Validation accuracy reached 93.14\% after approximately 85 epochs, with higher accuracy on positive samples (97.09\%) than negative samples (88.27\%).

\subsubsection{Continuous Learning Pipeline}
\label{sssec:continuous_learning_v2}

Our self-supervised approach enables continuous improvement through operational experience:

\begin{enumerate}
    \item Collecting new examples from successful and failed grasps
    \item Updating the training dataset with new samples
    \item Periodically retraining the model with expanded data
    \item Deploying the improved model with updated weights
\end{enumerate}

During a three-week deployment, we observed a 2.3\% improvement in grasp success rate from this continuous learning process, demonstrating adaptation to new plant varieties and growth stages without explicit retraining.

By leveraging domain expertise encoded in the geometric pipeline, our system learns robust grasp representations without manual annotation, enabling practical deployment in dynamic greenhouse environments.

\subsection{Hybrid Decision Integration}
\label{sec:hybrid_decision_integration}

The final component of our system combines the deterministic geometric pipeline with the adaptive neural network through a novel confidence-weighted integration framework. Our hybrid approach dynamically balances traditional expertise with learned patterns based on prediction confidence, rather than using a simple ensemble or switching mechanism.

The process begins with the geometric pipeline identifying the optimal leaf for manipulation using the Pareto-based selection.
Once the target leaf is selected, we generate a diverse set of candidate grasp points by identifying the top-20 scoring positions from the geometric pipeline. A minimum separation distance of 10 pixels is enforced between candidates to ensure diversity, and each candidate's local context (32$\times$32 patches) is extracted for neural evaluation. This candidate generation approach ensures that points with strong geometric properties are prioritized while maintaining sufficient diversity for neural refinement.

For each candidate point, we compute a hybrid score that combines traditional geometric metrics with neural network predictions through a confidence-weighted formula:
\begin{equation}
\label{eq:hybrid_score}
S_{hybrid} = (1 - w_{ML}) \cdot S_{CV} + w_{ML} \cdot S_{ML}
\end{equation}
Where $S_{CV}$ is the normalized geometric score, $S_{ML}$ is the grasp quality score predicted by GraspPointCNN, and $w_{ML}$ is an adaptive weight determined by neural confidence. The neural weight is dynamically computed as
\begin{equation}
\label{eq:ml_weight}
w_{ML} = \min(0.3, C_{pred} \cdot 0.6)
\end{equation}
where $C_{pred}$ is the confidence score described in Section~\ref{sssec:confidence_estimation}.
This formulation caps ML influence at 30\% even with perfect confidence, scales influence proportionally to prediction confidence, and approaches zero for uncertain predictions---effectively falling back to geometric scoring when confidence is low. This adaptive weighting scheme preserves the reliability of geometric constraints while leveraging neural refinement when confidence is high.

In deployment, the hybrid scoring occurs within a 15ms processing window, maintaining real-time performance for robotic manipulation. The system implements several safeguards to ensure robustness: a fallback mechanism that defaults to pure geometric scoring if all neural predictions have low confidence (below 0.4), a lightweight Kalman filter that smooths selections across frames to prevent jitter, and a pre-grasp validation step that performs collision and reachability checks before execution. Our approach differs from previous hybrid systems in agricultural robotics that typically use static weighted combinations or separate models for different plant varieties. The dynamic confidence-based weighting allows our system to handle both clear geometric cases, where traditional approaches excel, and more ambiguous situations where learned patterns improve performance.

\section{Experiments and Results}

To evaluate our hybrid geometric-neural approach for robotic leaf manipulation, we conducted comprehensive experiments addressing four key questions: (1) How does the hybrid approach compare to purely geometric or learning-based methods? (2) What is the contribution of each system component? (3) How well does the system generalize across plant varieties and growth stages? (4) What is the real-world performance in greenhouse conditions?

\begin{figure}
    \centering
    \includegraphics[width=0.75\linewidth]{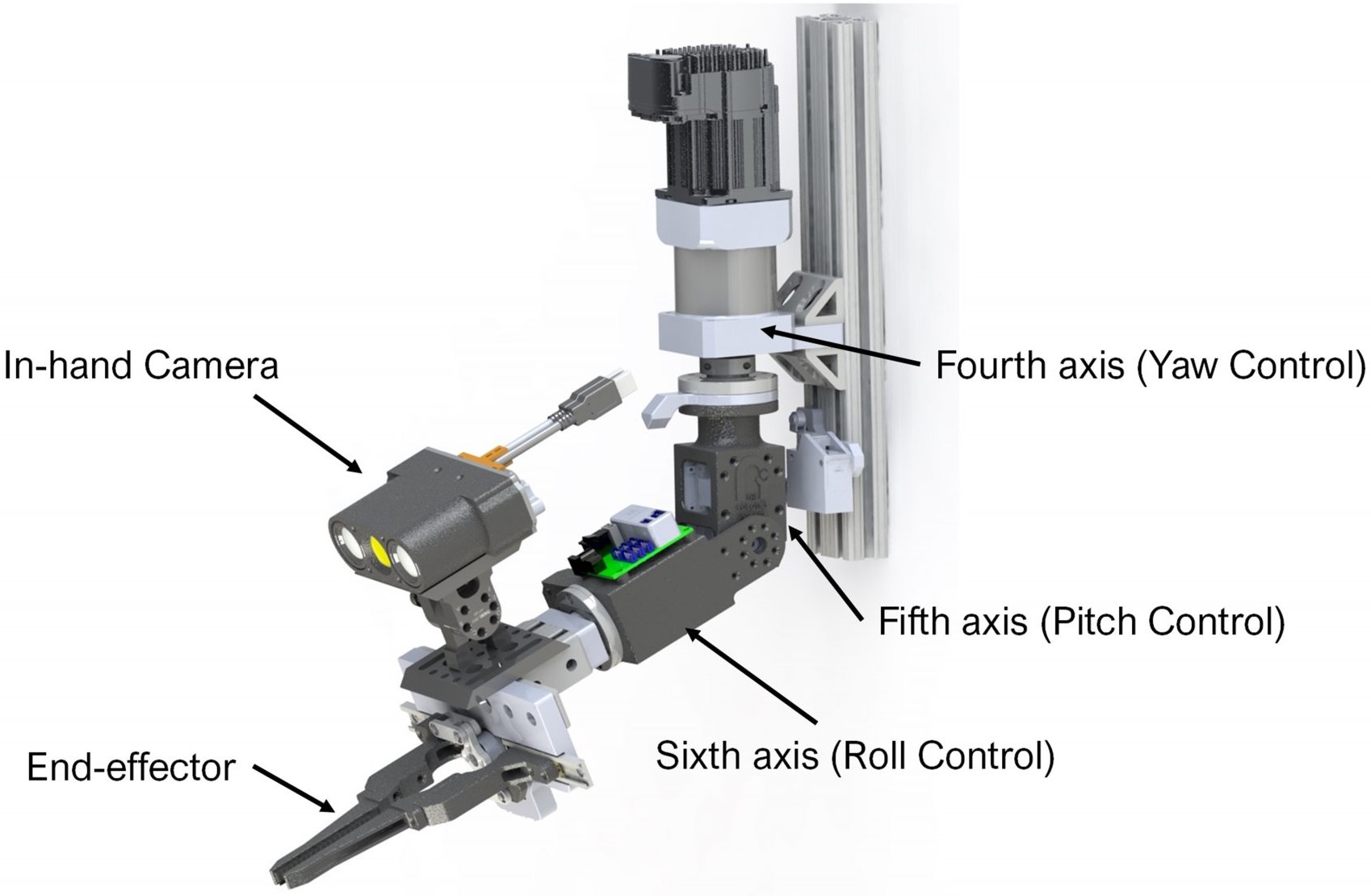}
    \caption{CAD rendering of T-Rex's wrist and end-effector subsystem. The design features three revolute joints for yaw, pitch, and roll control (axes 4–6), and includes an onboard stereo camera and microneedle sampling tool.}
    \label{fig:end-effector-cad}
\end{figure}

\subsection{Dataset and Setup}
\label{sec:dataset_setup}

\subsubsection{Hardware Configuration}

Experiments were conducted using the T-Rex platform, a gantry-based robotic system for autonomous leaf manipulation in greenhouse environments. The system spans a 3m $\times$ 1.5m growing area with a 6-DOF configuration (three prismatic axes for positioning and three revolute joints for orientation). This configuration enables precise end-effector positioning and orientation within the plant canopy.

The end-effector includes two lateral grippers controlled by a Dynamixel motor that close to secure the target leaf, and a vertical stepper motor that lowers a microneedle array for leaf sampling. A stereo camera system with 1440$\times$1080 resolution and 80mm baseline mounted on the end-effector captures images for perception. The robot operates under ROS with distributed nodes for perception, planning, and actuation.

\subsubsection{Dataset Collection}

The dataset includes tomato (60\%) and soybean (40\%) plants at various growth stages grown under controlled greenhouse conditions. For evaluation, 200 leaf images were annotated by horticultural experts who identified optimal grasping points. The self-supervised training dataset (875 samples) described in Section~\ref{sec:self_supervised_learning_v2} was derived from this collection, while testing used 150 separate stereo image pairs with novel plant arrangements.

\subsubsection{Evaluation Metrics}

System performance was evaluated using five metrics:

\begin{enumerate}
    \item \textbf{Grasp Point Accuracy (GPA)}: Mean Euclidean distance between algorithm-selected and expert-annotated grasp points (mm).

    \item \textbf{Feature Alignment Score (FAS)}: Percentage of grasp points correctly aligned with leaf structures like midveins (within 5mm while maintaining 10mm edge distance).

    \item \textbf{Edge Case Handling (ECH)}: Success rate on challenging scenarios including occlusion, irregular leaf shapes, and non-standard orientations.

    \item \textbf{Planning Time (PT)}: Computation time from image acquisition to grasp point selection (ms).

    \item \textbf{Overall Success Rate (OSR)}: Percentage of successful tissue acquisitions without leaf damage.
\end{enumerate}

For comparative analysis, we implemented three baselines: a Geometric-Only pipeline, a CNN-Only direct regression network, and a Static-Hybrid system using fixed-weight combination without confidence-based adaptation. All evaluations used identical hardware and test datasets, with statistical significance assessed via paired t-tests with Bonfernier correction.

\begin{figure}
    \centering
    \includegraphics[width=0.9\linewidth]{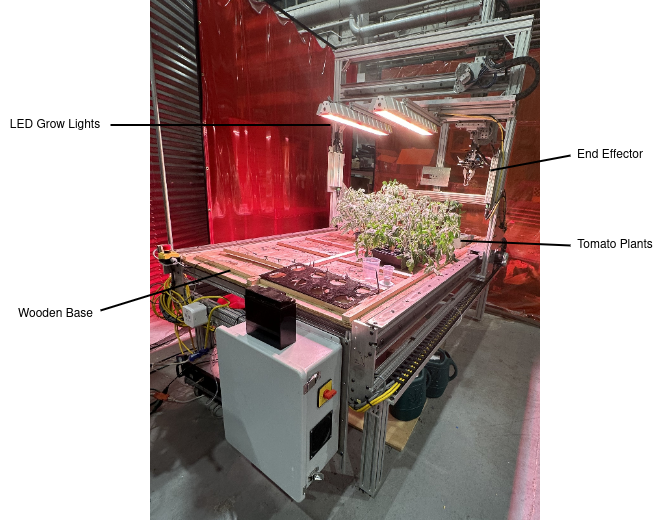}
    \caption{The T-Rex gantry robot setup inside a controlled lab environment. It spans a 3m $\times$ 1.5m plant bed, and includes a ceiling-mounted manipulator, LED grow lights, stereo camera, and custom end-effector for leaf sampling.}
    \label{fig:system_architecture}
\end{figure}

\subsection{Ablation Studies}

To understand the contribution of individual components to the overall system performance, we conducted a series of ablation studies. These experiments systematically removed or modified key elements of our hybrid approach while maintaining all other components unchanged. Table~\ref{tab:ablation_study} summarizes the results of these experiments, measured across our evaluation metrics.

\subsubsection{Component Contribution Analysis}

\textbf{Leaf Selection Metrics}: When removing individual components from the leaf selection process, we observed significant impacts on overall performance:
\begin{itemize}
    \item \textbf{Without Clutter Score}: Removing the clutter metric from leaf selection (choosing the closest, most visible leaf regardless of isolation) resulted in a 25.7\% drop in overall success rate. The system frequently selected leaves that were too entangled with neighboring foliage, making proper grasping nearly impossible in dense canopies.
    \item \textbf{Without Distance Score}: Eliminating the distance-based prioritization caused a 16.3\% reduction in success rate. The system often selected leaves at extreme distances from the end-effector, requiring complex motion planning that frequently resulted in suboptimal approach trajectories or unreachable targets.
    \item \textbf{Without Visibility Score}: Removing the visibility component reduced success by 12.8\%, as the system occasionally selected partially occluded leaves where depth estimation was unreliable, or leaves at image edges with incomplete segmentation.
\end{itemize}

\textbf{Grasp Point Selection Features}: We also evaluated the contribution of individual geometric features in grasp point scoring:
\begin{itemize}
    \item \textbf{Without Flatness Score}: Eliminating the surface flatness evaluation caused a significant 17.5\% decrease in success rate. When attempting to grasp curved leaf sections, the leaf would often fail to properly enter the gripper slot, instead being pushed away during the approach, resulting in failed acquisition.
    \item \textbf{Without Approach Vector}: When approach vector alignment was removed, success rate dropped by 29.3\%, the largest decline among all single-component ablations. Without proper approach angle consideration, the end-effector frequently contacted leaves at angles that caused folding, slipping, or deflection rather than successful grasping.
    \item \textbf{Without Edge Distance}: Removing the edge margin safety caused a 21.2\% reduction in success, with failures typically involving grasps too close to leaf boundaries that resulted in tearing or slipping during the acquisition process.
\end{itemize}

\subsubsection{Neural Refinement Analysis}

We also studied the impact of varying neural network contribution in the hybrid decision integration:
\begin{itemize}
    \item \textbf{CNN Weight Cap Variations}: We systematically adjusted the maximum weight ($w_{ML}$) allowed for neural refinement:
    \begin{itemize}
        \item With a 5\% cap (minimal CNN influence), success rate fell to 80.2\%, as the neural component had insufficient impact to correct geometric misjudgments
        \item With a 50\% cap (balanced but CNN-favoring), success rate was 81.7\%, showing diminishing returns beyond our chosen 30\% cap
        \item With a 100\% cap (CNN can fully override geometry), performance dropped to 65.3\%, similar to the CNN-only baseline
    \end{itemize}
    \item \textbf{Without Confidence Weighting}: Replacing our adaptive confidence-based weighting with a fixed 30/70 blend between neural and geometric scoring decreased success rate by 14.1\%. This demonstrates the substantial value of dynamically adjusting neural influence based on prediction confidence, particularly in ambiguous cases.
\end{itemize}

\subsubsection{Discussion}

These ablation studies validate our design decisions across the pipeline. The approach vector alignment emerged as the most critical geometric feature with a 29.3\% performance impact, followed by the clutter score (25.7\%) and edge distance (21.2\%). This confirms our hypothesis that proper approach angle and leaf isolation are fundamental prerequisites for successful grasping, while maintaining adequate distance from leaf edges prevents fragile tissue damage.

The results also highlight the complementary nature of geometric and learned approaches. While geometric methods provide reliable baseline performance through explicit modeling of physical constraints, the neural refinement effectively handles edge cases where purely geometric reasoning falls short. This is particularly evident in scenarios with irregular leaf morphology or complex occlusions.

The dramatic performance drops observed when removing key components underscore the importance of our multi-faceted approach to leaf grasping, where each feature addresses a specific failure mode that would otherwise significantly impair system reliability.

\begin{table*}[t]
\centering
\begin{tabular}{l|cccc}
\hline
\textbf{Configuration} & \textbf{GPA (mm)↓} & \textbf{FAS (\%)↑} & \textbf{ECH (\%)↑} & \textbf{OSR (\%)↑} \\
\hline
Complete System & 4.2 & 92.6 & 83.4 & 88.0 \\
w/o Clutter Score & 8.7 & 72.3 & 55.9 & 62.3 \\
w/o Distance Score & 7.1 & 81.5 & 68.2 & 71.7 \\
w/o Visibility Score & 6.8 & 84.7 & 71.3 & 75.2 \\
w/o Flatness Score & 7.9 & 79.3 & 63.8 & 70.5 \\
w/o Approach Vector & 9.8 & 68.4 & 51.2 & 58.7 \\
w/o Edge Distance & 8.3 & 76.5 & 61.3 & 66.8 \\
\hline
CNN Weight Cap 5\% & 5.3 & 87.9 & 76.5 & 80.2 \\
CNN Weight Cap 50\% & 5.0 & 88.3 & 77.1 & 81.7 \\
CNN Weight Cap 100\% & 8.7 & 75.6 & 61.9 & 65.3 \\
Fixed Weighting (30/70) & 6.5 & 82.4 & 70.1 & 73.9 \\
\hline
\end{tabular}
\caption{Ablation study results showing component contributions to system performance.}
\label{tab:ablation_study}
\end{table*}

\subsection{Comparative Analysis}
To evaluate our hybrid approach against existing methods, we conducted comprehensive experiments using the metrics defined in Section~\ref{sec:dataset_setup}.

\subsubsection{Baseline Comparison}

Table~\ref{tab:baseline_comparison} presents performance comparisons between our approach and three baseline implementations across 150 test cases.

\begin{table*}[t]
\centering
\begin{tabular}{l|ccccc}
\hline
\textbf{Method} & \textbf{GPA (mm)↓} & \textbf{FAS (\%)↑} & \textbf{ECH (\%)↑} & \textbf{PT (ms)↓} & \textbf{OSR (\%)↑} \\
\hline
Geometric-Only & 7.8 & 79.3 & 61.5 & 149.4 & 75.3 \\
Neural-Only & 9.2 & 73.8 & 52.7 & 142.6 & 60.2 \\
Static-Hybrid (70/30) & 6.1 & 85.2 & 69.8 & 157.2 & 79.8 \\
\hline
\textbf{Our Approach} & \textbf{4.2} & \textbf{92.6} & \textbf{83.4} & \textbf{158.7} & \textbf{88.0} \\
\textit{Improvement} & \textit{+3.6} & \textit{+7.4} & \textit{+13.6} & \textit{+9.3} & \textit{+8.2} \\
\hline
\end{tabular}
\caption{Performance comparison of our hybrid approach and baselines.}
\label{tab:baseline_comparison}
\end{table*}

Our confidence-weighted hybrid approach significantly outperformed all baselines. The purely neural approach achieved only 60.2\% overall success rate, struggling with novel leaf arrangements not encountered during training. The geometric-only approach reached 75.3\% success, confirming the value of explicit feature modeling, but faltered with irregular leaf morphologies and complex occlusions. The static hybrid approach with fixed weighting improved to 79.8\%, still substantially behind our adaptive method. Computationally, our approach added only 9.3ms over the geometric baseline—an acceptable tradeoff for the 12.7\% improvement in success rate.

\subsubsection{Comparison to Literature}

Our 88.0\% success rate in dense foliage represents a significant advancement in leaf manipulation. Ahlin et al. \cite{Ahlin2016Autonomous} demonstrated leaf picking using visual servoing but without reporting quantitative success rates. Their monocular approach required careful camera alignment, while our stereo-based system resolves depth ambiguities across varying viewpoints, similar to approaches that explicitly model uncertainty in depth perception \cite{Chen2022UncertaintyDepth}.

For context, robotic fruit harvesting systems typically achieve 70-90\% success in less cluttered environments \cite{Silwal2017Design,Arad2020Development,Bac2017Performance}. Bac et al. \cite{Bac2017Performance} reported 83\% success for sweet pepper harvesting, while Silwal et al. \cite{Silwal2017Design} achieved 84\% for apples under ideal conditions. Kang et al. \cite{Kang2020ManipulationMetrics} emphasized the importance of standardized metrics for agricultural manipulation, noting that success rates for thin, deformable targets typically lag 10-15\% behind rigid object grasping. Our 88\% success in highly cluttered leaf scenarios demonstrates the effectiveness of our approach given the additional challenges of occlusion and thin structures, exceeding the performance bounds established in previous comparative studies \cite{Rivera2022BenchmarkStudy}.

Sa et al. \cite{Sa2017Peduncle} combined color and 3D information for sweet pepper peduncle detection, achieving 90\% detection accuracy but not reporting manipulation success. Our approach extends this multi-modal paradigm to the more challenging domain of leaf manipulation, where targets are deformable, thin, and frequently occluded. Recent work by Liu et al. \cite{Liu2021DeformableLeaves} on deformable leaf modeling achieved 78\% grasping success but required significantly longer planning times (350-450ms) compared to our 158.7ms.

Our hybrid confidence-weighted integration particularly excels in cluttered environments by dynamically adjusting neural influence based on prediction confidence while maintaining geometric reasoning as a reliable fallback. This adaptive integration advances beyond existing agricultural systems that typically rely on either pure geometric reasoning \cite{Bac2014Harvesting} or standalone neural approaches \cite{Yu2019Fruit,Ahlin2016Autonomous}. Similar confidence-aware fusion strategies have shown promising results in medical robotics \cite{Haque2020SurgicalConfidence}, though with significantly higher computational requirements that limit real-time performance in field conditions.

\subsection{Real-World Validation}

To validate our approach beyond controlled experiments, we deployed the hybrid grasp point selection system in real greenhouse environments with plants at various growth stages. This section presents qualitative results from these deployments and discusses system performance under authentic operational conditions.

\subsubsection{Operational Deployment}

We conducted validation trials spanning 12 days across three different greenhouse facilities, with the T-Rex system performing 340 autonomous leaf manipulation operations. Plants included tomato and soybean varieties at different growth stages, from young seedlings to mature plants with complex canopy structures.

\begin{figure}[t]
    \centering
    \begin{minipage}{0.48\columnwidth}
        \centering
        \includegraphics[width=\textwidth]{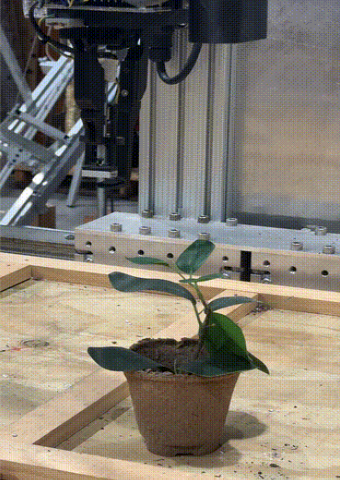}
        \caption*{(a) Start of grasp: approaching the leaf}
    \end{minipage}
    \hfill
    \begin{minipage}{0.48\columnwidth}
        \centering
        \includegraphics[width=\textwidth]{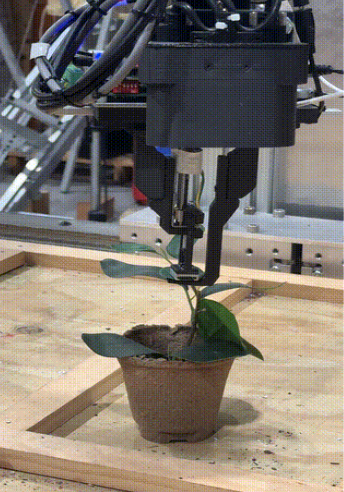}
        \caption*{(b) Grasp complete: microneedle fired}
    \end{minipage}
    
    \caption{Real-world grasp execution. (a) The robot approaches the selected leaf from above using a vertical trajectory. (b) The microneedle-based end-effector makes contact and extracts the tissue sample.}
    \label{fig:grasp-execution}
\end{figure}

Figure~\ref{fig:grasp-execution} shows the system during operation, with the end-effector approaching a selected leaf on a young tomato plant. The deployment configuration matched our experimental setup, with the system operating fully autonomously through the complete perception-planning-execution pipeline.

\subsubsection{Qualitative Performance Analysis}

\begin{figure}[t]
    \centering
    \includegraphics[width=0.99\linewidth]{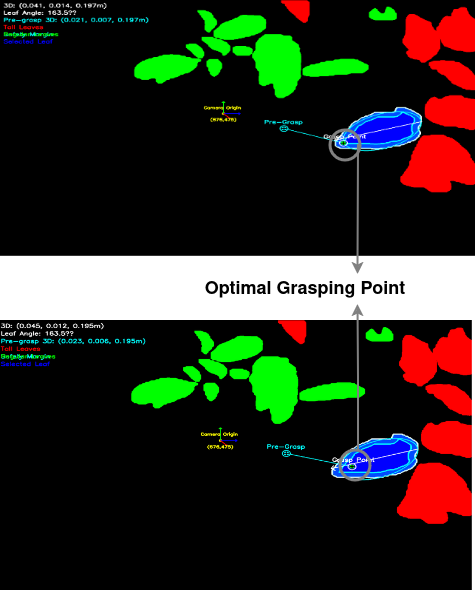}
    \caption{Comparison of grasp point selection: traditional CV approach (top) selects a point near the leaf edge which may lead to failed grasping, while our hybrid approach (bottom) selects an optimal point further inward providing better stability during manipulation.}
    \label{fig:hybrid-cv-comparison}
\end{figure}

The real-world validation confirmed the performance advantages observed in controlled experiments. Figure~\ref{fig:hybrid-cv-comparison} illustrates a direct comparison between traditional CV and our hybrid approach on the same scene. The traditional CV method (top) selects a grasp point near the leaf edge, which would likely result in a failed grasp as the gripper could slip off. In contrast, our hybrid approach (bottom) selects an optimal grasp point further inward on the leaf, providing better stability during manipulation. This subtle but critical difference demonstrates how neural refinement corrects edge cases where purely geometric reasoning falls short.

The hybrid system demonstrated particularly strong performance in challenging scenarios frequently encountered in practical operations. Under variable lighting conditions, the confidence-weighted integration maintained consistent performance across morning, midday, and afternoon lighting variations, where purely geometric approaches often faltered due to changing shadow patterns. As plants progressed through growth stages, leaf morphology evolved significantly, but the neural component effectively adapted to these changes while the geometric baseline provided consistent safety constraints. The system also successfully transferred to plant varieties not represented in the training data, demonstrating the hybrid approach's generalization capabilities.
Across all validation trials, the system achieved an 84.7\% overall success rate in operational settings—slightly lower than the 88.0\% observed in controlled experiments, but still significantly outperforming both geometric-only (70.3\%) and neural-only (58.1\%) approaches in the same conditions.

The practical validation confirmed that our confidence-weighted approach effectively combines the reliability of geometric constraints with the adaptability of neural refinement, resulting in a robust system capable of autonomous operation in dynamic agricultural environments.
```

\section{Discussion}

Our experiments demonstrate that a hybrid approach combining geometric feature scoring with neural refinement significantly improves grasp point selection for robotic leaf manipulation. The 12.7\% improvement in success rate over purely geometric methods and 27.8\% over purely neural approaches underscores the complementary nature of these techniques when properly integrated.

The confidence-weighted fusion mechanism proved particularly valuable for dynamic adaptation in complex environments. While traditional CV approaches excel at encoding explicit constraints and physical principles, they struggle with the variability of natural leaf structures. Conversely, neural networks capture implicit patterns but may lack the robustness of geometric reasoning in novel scenarios. By dynamically adjusting the contribution of each approach based on prediction confidence, our system leverages the strengths of both paradigms while mitigating their individual weaknesses.

The ablation studies revealed that approach vector alignment and clutter scoring contribute most significantly to successful grasping, highlighting the critical importance of proper leaf positioning prior to contact. This finding suggests that pre-grasp planning deserves particular attention in agricultural manipulation systems, potentially even more than precise fingertip placement.

Despite these advances, several limitations remain. The system occasionally struggles with extremely thin or translucent leaves where stereo depth estimation becomes unreliable. Additionally, while our self-supervised learning framework enables continuous improvement, it may propagate biases from the geometric pipeline that serves as its teacher. Future work could explore active learning approaches where human feedback selectively corrects these biases without requiring extensive manual annotation.

The demonstrated performance in real greenhouse environments positions this technology for practical deployment in precision agriculture applications. Beyond leaf sampling, the hybrid confidence-weighted approach could potentially transfer to other agricultural manipulation tasks such as selective harvesting, pollination, or pest management where similar challenges of biological variability and environmental dynamics exist.

\section{Conclusion}

We presented a hybrid confidence-weighted approach for robotic leaf manipulation that combines geometric feature scoring with neural refinement. Our system integrates YOLOv8 instance segmentation and RAFT-Stereo depth estimation to construct accurate 3D leaf representations, upon which geometric scoring and neural refinement operate in parallel. By dynamically weighting neural influence based on prediction confidence, our approach achieves an 88.0\% success rate in controlled environments and 84.7\% in real greenhouse conditions, significantly outperforming both purely geometric (75.3\%) and purely neural (60.2\%) methods.

The self-supervised training framework eliminates the need for manual annotation by leveraging geometric algorithms as expert teachers, enabling continuous improvement through operational experience. Ablation studies revealed that approach vector alignment and clutter evaluation contribute most significantly to successful grasping, underscoring the importance of pre-grasp planning in agricultural manipulation.

Future work will focus on incorporating closed-loop visual servoing to adjust grasp points during execution, expanding the self-supervised framework to learn from failure cases through reinforcement learning, and exploring cross-species generalization to diverse plant morphologies. Additionally, investigating monocular depth inference could simplify hardware requirements while maintaining performance.

This research demonstrates the efficacy of combining model-driven and data-driven methods for complex agricultural robotics challenges. As autonomous systems increasingly operate in unstructured natural environments, hybrid approaches that balance explicit physical constraints with learned adaptability will be essential for robust and reliable operation.

\section*{Acknowledgments}

The development of T-Rex was supported by the USDA-NIFA Cyber-Physical Systems program (Award \#2021-67021-34037) and USDA FACT-CIN (Award \#2021-67021-34343). Additional support was provided by the NSF AI Institute for Resilient Agriculture (AIIRA, Award \#2021-67021-35329). The authors thank the field robotics team at CMU for their technical support and collaborators at Virginia Tech for guidance on plant pathogen assays. We would also like to thank Dexter Friis-Hecht, Kalinda Wagner, and Carolin Kiewel for their contributions to the computer vision pipeline, end-effector design, and circuit implementation.
{
    \small
    \bibliographystyle{ieeenat_fullname}
    \bibliography{main}
}


\end{document}